\documentclass[11pt]{article}
\usepackage{eacl2017}
\usepackage{times}
\usepackage{url}
\usepackage{latexsym}
\usepackage{graphicx}
\usepackage{booktabs}

\usepackage{tabularx,arydshln}
\usepackage{color,soul}

\setul{0.5ex}{0.3ex}
\setulcolor{blue}

\eaclfinalcopy

\title{Leveraging Sentiment for Offensive Text Classification}

\author{
  Khondoker Ittehadul Islam \\
  University of Groningen \\
  {\tt k.i.islam@student.rug.nl}\\
  \\
}

\date{}

\begin{document}
\maketitle
\begin{abstract}
In this paper, we conduct experiment to analyze whether models can classify offensive texts better with the help of sentiment. We conduct this experiment on the SemEval 2019 task 6, OLID, dataset. First, we utilize pre-trained language models to predict the sentiment of each instance. Later we pick the model that achieved the best performance on the OLID test set, and train it on the augmented OLID set to analyze the performance. Results show that utilizing sentiment increases the overall performance of the model.{\footnote{Codes released at \url{https://github.com/KhondokerIslam/senti-affects-olid}}}

\end{abstract}

\section{Introduction}

Recent years witnessed the upsurge of user generated content which requires proper regulation to prevent the spread of hate speech causing through offensive language \cite{cheng2017anyone}.
Automatically classifying user-generated text to classify offensive text is a difficult task due to sarcasm and irony \cite{sharma2021deep}.

In SemEval-2019 Task 6: Identifying and Categorizing Offensive Language in Social Media
\cite{zampieri2019semeval}, the organizers collected
English tweets through Twitter API and annotated them hierarchically regarding offensive language. A wide range of teams participated in the competition itself while this dataset, OLID, is also being currently utilized to surge research to better classify instances as offensive or non-offensive.

On the other hand, sentiment analysis is a task that is closely associated with offensive text classification \cite{sharma2021deep} as sentiment drives certain people to use offensive language.
Even though there is a huge amount of work conducted on predicting sentiment from English text to the best of our knowledge, no one explored prepending predicted sentiment with instances to classify offensive language.

Furthermore, with the advent of popular deep learning models, we analyze whether offensive text can be better classified with the help of sentiment prepending on the text. We summarize our contribution as follows:

\begin{itemize}
    \item We demonstrate the sentiment distribution on the OLID dataset after predicting the corresponding sentiment with a pre-trained language model. We notice a huge prevalence of \textit{Neutral} sentiment.
    
    \item We experiment with different techniques such as linguistic features, recurrent neural networks, and pre-trained language models on the OLID dataset with and without sentiment; and show that models that utilize sentiment outperform all other models.

    \item We report our error analysis with and without leveraging sentiment on the best performing model to help craft a better research path for future researchers.

\end{itemize}

\section{Related Work}

The dataset we use to experiment is the OLID dataset from SemEval Task 1 competition \cite{zampieri2019semeval}.
The data were English tweets scrapped from social media site, Twitter with each tweets corresponds to two labels OFF and NOT.
115 teams participated on the competition, out which 13\% used LSTM, Bi-LSTM model and only 8\% used BERT models showcasing the general acceptability of these models which prompted us to explore these models capabilities on our research question. 
Furthermore, among the small BERT model users, seven team ranked on the top 10 showcasing the natural ability of BERT model to perform better on this task.

The dataset has class imbalance phenomenon resulting one class to detect more often than the others which consequently effects the competition evaluated metric, macro-F1 average. 
This phenomenon was only tackled by \newcite{nikolov2019nikolov} applying several techniques and achieved 2nd place on this subtask. Out of three techniques experimented by this team, we only utilized their technique of assigning lower weights to classes that are most labeled while assigning higher weights to class which are less labeled. Furthermore, \newcite{anwar2020tac} utilized focal loss \cite{ross2017focal} to overcome the class imbalance phenomenon on the OLID task. Similiar to class weights, focal loss also down weights well classified labels and gives importance to the difficult to classify labels. The authors of focal loss suggests $\alpha$ and $\gamma$ value to be set to $1$ and $2$ respectively to attain better results. As a result we utilize the focal loss on the transformer setting similar to the work of \newcite{anwar2020tac}. 

Previous works on utilizing sentiment to classify offensive text mostly comprised considering sentiment as a separate feature and concatenating their vectors to analyze their effect on this task \cite{kannan2019tukast,ghosh2020iitp,bollegala2011using,peng2020knowing}.
However, since the advent of Transformer models where appending or prepending sentiment as a text to the initial data, explored by only \cite{althobaiti2022bert}. They appended the sentiment to their Arabic tweet and analyzed the performance of their monolingual BERT model with and without sentiment. Our research work is similar to theirs except that they did their analysis on Arabic tweets and we will conduct them on English tweets. Furthermore, we prepend the predicted sentiment instead of the appending technique, as this will provide more clarity. However, differentiating the performance of models between prepending or appending is beyond the scope of our research work.

\section{Data}

In this section, we present the data statistics of OLID and data preparation of our prepend predicted sentiment \textit{PPS} on OLID, i.e., \textit{PPS}-OLID.

\subsection{OLID}

The task dataset consists of 14,100 tweets, each label with either OFF (i.e., offensive) or NOT (i.e., not offensive), with two annotators agreeing 60\% of the time. The average sentence length is $23.86$ words with token sizes ranging from 3 to 104. The labels, OFF and NOT, are distributed with $32.90\%$ (4,640 instances) and $67.10\%$ (9,460 instances), respectively, indicating the label imbalance nature of the dataset.

\subsection{\textit{PPS}-OLID}

For the sentiment prepending task, we considered DeBERTa V3 \cite{he2020deberta} as it outperforms all existing models on the SST dataset. However, with the OLID dataset consisting of user-generated Twitter data, we looked into Hugging Face\footnote{https://huggingface.co/} for a model which further fine-tuned DeBERTa V3 on a similar dataset task as conducting such a step is beyond the scope of this work. Furthermore, we were also looking for models that could output three sentiment fine-grained outputs, i.e., neutral, negative, and positive with respect to SST labeling two sentiments positive and negative. This is because non-offensive tweets tend to have a huge correlation with neutral sentiment \cite{sharma2021deep}. As a result, based on the most downloads, we select DeBERTa-v3-small-ft-senti \cite{manuel_romero_2024} which demonstrates 99.40 f1 scores after fine-tuning on various English-sentiment datasets.
Consequently, for each instance of OLID, we predict the sentiment and prepend on the instance. We will refer to this modification of OLID as \textit{PPS}-OLID throughout the paper for clarification. The sample of \textit{PPS}-OLID is provided in Table \ref{tab:sample_with_append}.

\begin{table}[t]
    \centering
    \resizebox{0.9\columnwidth}{!}{%
    \begin{tabular}{l  c}
        \toprule
        \textbf{Label} & \textbf{Text} \\
        \midrule
        
         OFF & \ul{negative} @USER The Liberals are mentally unstable!! \\

        \bottomrule
    \end{tabular}}
    \caption{Dataset sample of \textit{PPS}-OLID. The prepend predicted sentiment is underlined with blue color.}
    \label{tab:sample_with_append}
\end{table}

\begin{figure}
    % \centering
        \includegraphics[width=1.02\columnwidth]{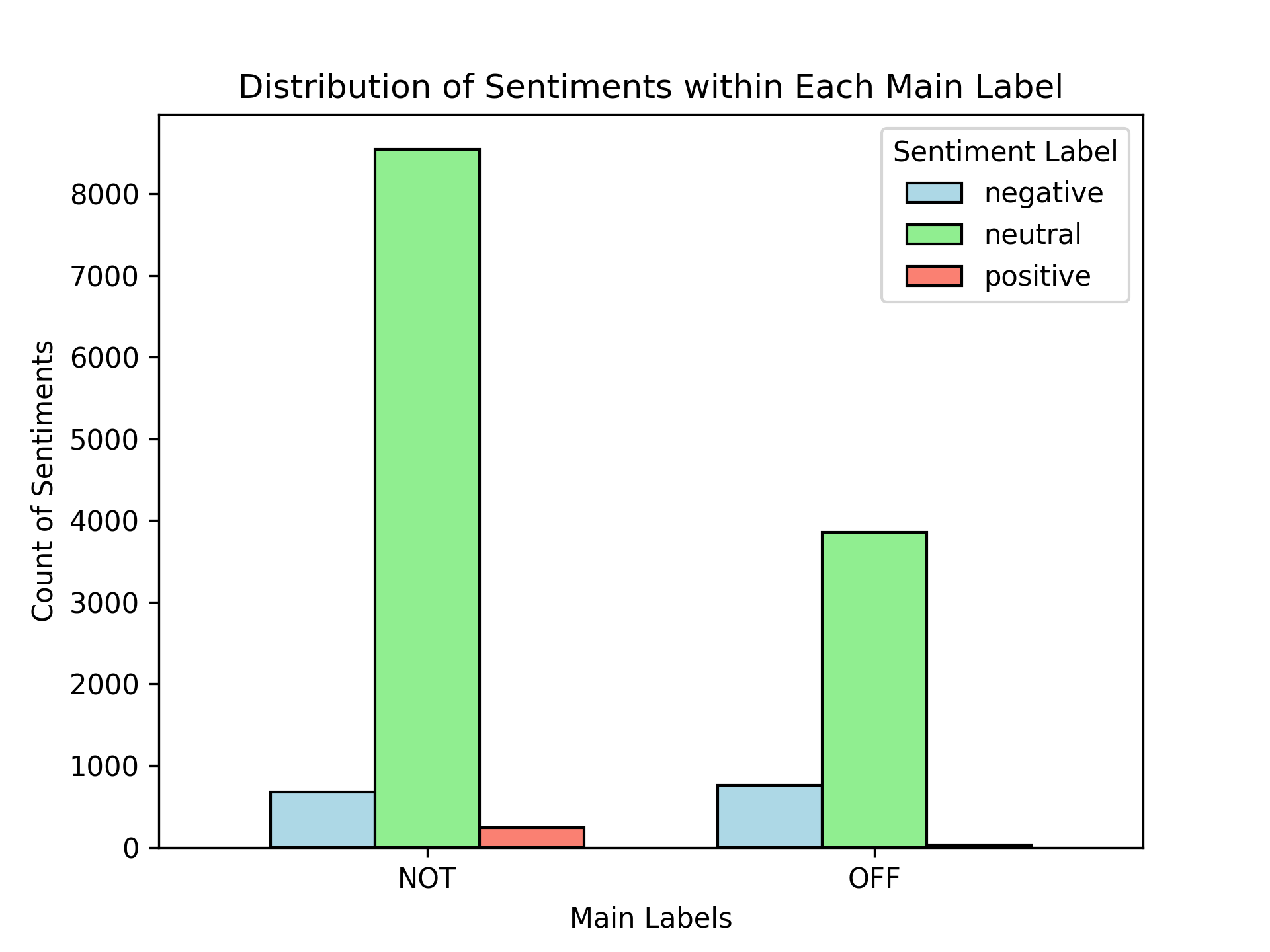}
    \caption{Distribution of \textit{PPS} per OLID labels.}
    \label{fig:sentiment_dist_per_label}
\end{figure}

Figure \ref{fig:sentiment_dist_per_label} demonstrates the distribution of sentiment on \textit{PPS}-OLID. Notice the huge presence of \textit{Neutral} sentiment across the dataset with NOT label comprising of the most \textit{Neutral} sentiment. Furthermore, there is no \textit{Positive} sentiment on the OFF instances while \textit{Negative} sentiment are more present on OFF label then NOT label. These are intuitive as offensive labels are not expected to have positive sentiment while offensive labels are most likely to have more negative sentiment compared to the non-offensive comments.

\section{Methodology}

In this section, we describe the methods used to investigate the performance on the OLID task.

\subsection{Lexical Feature} 

We extract word (1-4) and character (1-4) n-grams from the instances. Each instance is then vectorized using TF-IDF weighted scores and passed to the linear SVM models \cite{cortes1995support} as these models optimally create decision boundary between feature vectors of relevance and novelty when selecting features for relevance-based feature selection task \cite{lee2006information,trstenjak2014knn} using the C-value which controls the trade-off between the margin and misclassification.

\subsection{Recurrent neural network}

Neural networks, a system of interconnected nodes called neurons, represent a set of algorithms that intend to mimic the human brain’s work for identifying hidden relationships within a set of data. In multi-layered neural networks, the neurons are stacked into interconnected layers called hidden layers. The input layers collect the signals and pass them through multiple layers down to the final output layer, which maps the
inputs to the classification categories \cite{sharma2021deep}. 

One such architecture that uses neural networks is The long short-term memory (LSTM) \cite{hochreiter1997long}. The architecture is built in a way that effectively learns which information might be needed later on in a sequence and when that information is no longer needed \cite{gers2000learning}. 
However, LSTM models have the capability to suffer from the exploding gradient problem, i.e., when large error gradients accumulate and result in very large updates to neural network model weights during training \cite{mishra2024recurrent}.

We use an LSTM network that encodes text from the forward directions and creates a 1D vector. We compute the weighted sum of the vectors and predict the offensive polarity through an output layer. We also experiment with a bidirectional long short-term memory (Bi-LSTM) network that additionally encodes text in backward directions and creates a 2D vector for each direction.

For the embedding layer, we used GloVe embeddings trained on the Twitter corpus \cite{pennington2014glove}. We used 200-dimensional embeddings to represent the tweets. The embeddings of the tweets were passed through a dropout layer to add regularization to our model and help improve learning.

\subsection{Pre-trained Language Model}

In past years, large pre-trained language models
like BERT \cite{devlin2018bert} have shown impressive performance on a wide range of text classification tasks due to breaking words into subwords and randomly masking them, allowing the model to develop an understanding of the surrounding context. This approach enables BERT to capture the context of long convoluted sentences, a natural phenomenon of tweets. 

Due to BERT's existing powerful performance on OLID tasks, we only probe into the latest, computationally powerful LLM model, DeBERTa \cite{he2020deberta}. DeBERTa retains the masking capability of BERT while introducing two vectors for each word to better influence attention weights based on their positions in the sentence. They perform it by replacing mask tokens with incorrect tokens. The latest release DeBERTa v3 had just the layers and hidden upgrades. We experiment with both DeBERTa and DeBERTa v3 to analyze their results on this task. For these models, we consider the problem as a multi-class classification task and choose the class that had the maximum value on the logits.

\subsection{Regularization}

Regularization is a process that prevents networks from over-fitting. Below, we provide details of some forms we utilized on our task.

\paragraph{Optimizer} We use Adam and AdamW \cite{kingma2014adam} optimization algorithms that is designed for training deep neural network with an adaptive learning rate.

\paragraph{Loss Function} This function updates the weights of the neurons of a neural network by calculating the mistake errors during backpropagation. As OLID task a two-class problem, we used Binary cross-entropy or sigmoid cross-entropy loss. It predicts the
probability of a sample belonging to one of the classes \cite{shore1981properties}.

\paragraph{Activation Function} In a neural network we apply an activation function to get the current layer's output, which then becomes the input of the next layer \cite{ramachandran2017searching}. This functions helps map complex information to the corresponding value indicated information such as sigmoid or Relu. 

\paragraph{L2} This regularization form prevent the weights from obtaining large values that prevent significant variances in weights, improving the
model \cite{cortes2012l2}. We utilize this form of regularization when building LSTM models.

\paragraph{Dropout} This form drops out neurons of a network randomly. The value initialized is a likelihood parameter to drop random units \cite{hinton2012improving}. We utilize it between hidden and final layers of LSTM.

\paragraph{Random seed} Pseudorandom number generator to initialize a vector in order to keep the results same on across multiple re-runs.

\section{ Experimental Setup }

We implement our experimental framework using Scikit-learn \cite{pedregosa2011scikit}, Tensorflow \cite{abadi2016tensorflow}, and Transformers \cite{wolf-etal-2020-transformers}. We evaluate our methods using macro averaged F1. Below, we present our experimental settings.

\subsection{Preprocessing}
We convert all the cases to lowercase. We replace user mention '@User' to '$<$user$>$'. We remove the \# symbol and only keep the text. Later, we remove the repeated characters of words. All such using python $re$ library by \newcite{van1995python}.

\subsection{Lexical Feature}

We only tune the regularizer C\footnote{We tested on these values: $1e^{-3}, 1e^{-2}, 0.1, 1$ (best), $10$.} of the SVM model. 

\subsection{LSTM}
We set the random seed of $np$, $tf$ to $1234$. To handle class imbalance, we set the class weight as $balanced$. We used Adam optimizer with clipnormal set to $1.0$. For this architecture, we considered this problem as binary classification task, as a result, we used Binary cross-entropy as the loss function, sigmoid as the activation of the final layer with threshold of $0.5$ differentiating between the labels classification. However, for the activation between the layers, we consider Relu. Furthermore, we used L2 regularization with value set to $0.01$. We monitor the validation loss during training using ReduceLROnPlateau\footnote{\url{https://www.tensorflow.org/api_docs/python/tf/keras/callbacks/ReduceLROnPlateau}} with factor set to $0.1$ and minimum learning rate set to 1 $\times$ 10$^{-6}$. We utilize EarlyStopping\footnote{\url{https://keras.io/api/callbacks/early_stopping/}} with patience set to $7$ and restoring best weights. We also experiment with learning rate, layers, units, dropout, batch size, epochs and present them in Table \ref{tab:Hyper_Parameter_LSTM_Tab} at Appendix \ref{app:app_sec_hyper_paramete}.

\subsection{Transformers Model}
We set the random seeds of $np$, $torch$ to $42$ during DeBERTa model experiments. We also experiment with different maximum length with tokenizer initialized as AutoTokenizer\footnote{\url{https://huggingface.co/docs/transformers/v4.46.0/en/model_doc/auto}} with padding and truncation. For the optimizer, we choose AdamW. To handle class imbalance, we use Focal loss with $\alpha$ and $\gamma$ set to $1$ and $2$ respectively. We also experiment with different batch size, learning rate, weight decay, epoch and report them in Table \ref{tab:Hyper_Parameter_Transformer_Tab} at Appendix \ref{app:app_sec_hyper_paramete}.

\section{Results}

We report our experimental results on the test set in Table \ref{tab:test_res}. 
Among the word n-grams, we observe better performance with unigram $50.45$ compared to bigram ($42.59$) and trigram ($25.21$).
Combining bigram with unigram lifts the unigram F1 by $4.23$ (i.e., $54.68$). Furthermore, adding trigram to that combination increases the rate of improvement by $1.08$, and we achieve $55.71$ F1.
We observe a similar classification performance with the character n-grams.

While character $3$, and $4$ grams' performances are around $16$-$17$\% higher than character bigram, the difference among their F1 scores is low.
Furthermore, different combinations of the character n-grams do show relatively higher gains ($0.55$) over the $3$-gram.
Combining all character n-grams yields a small gain of $0.02$ over the most robust character combination of $2$-$4$-gram feature.
However, we do not observe any significant shift in the precision and recall scores for character n-gram combinations.
This implies that the task highly depends on word units and does not rely much on the subword level information.

On the neural model results, we notice a significant contrast between the LSTM model and the Bi-LSTM model ($46.58$ vs $76.78$). 
Even though the LSTM model has a higher recall, the F1 score of this model is below the linguistic feature combinations.
One reason for this poor performance could be the exploding gradient problem as the errors to work on this problem were too much for this base model whereas it's bidirectional capabilities were able to learn and understand the context better. Having said that, giving a valid reason for this performance is beyond the scope of our work.

Among the transformer models, we notice an increase in score across precision, recall, and F1 on the DeBERTa model. However, with DeBERTa v3 the performance on all three metrics drops, especially of the F1-score by $1.06$. This is surprising as DeBERTa v3 is just a computational upgrade of its previous version. One reason could be it replacing masks with false tokens on higher-layer architecture was unable to capture the complex nature of the offensive tweets. Future work could revolve around this path.

Finally, when we feed the best model on the OLID task, DeBERTa, on our \textit{PPS}-OLID dataset, we find an increase of F1-score by $0.28$ from DeBERTa+\textit{PPS}. Even though there is a drop in recall, we also notice an increase in precision by $0.24$, concluding that prepending predicted sentiment is beneficial for offensive text classification tasks.

\begin{table}[t]
    \centering
    \resizebox{0.9\columnwidth}{!}{%
    \begin{tabular}{l c c c}
        \toprule
        \textbf{Method} & \textbf{Precision} &\textbf{Recall} & \textbf{F1} \\
        \midrule
        
         Unigram (U) & 54.90 &46.67 & 50.45\\
          Bigram (B) &	47.92 &  38.33 & 42.59  \\
        Trigram (T) &  40.37 & 18.33   &	25.21 \\	
         U + B & 66.80 &46.257 & 54.68\\
         U + B + T & 61.62 &50.83 & 55.71\\
         
         %  \hline
         Char 2-gram (C2) & 52.23&	34.17&	41.31\\
         Char 3-gram (C3) & 65.96 &51.67 & 57.94\\
         Char 4-gram (C4) & 67.05&	49.17&	56.73\\
         C2 + C3 + C4 & 67.39 & 51.67 & 58.49\\
         C1 + C2 + C3 + C4 & 68.93& 50.83 & 58.51\\

          \hdashline
         LSTM & 66.42 &	52.90 &	46.58\\
         Bi-LSTM & 76.39 &77.36 & 76.78\\

          \hdashline
         DeBERTa & 82.94 &	81.71 &	82.28\\
         DeBERTa v3 &  82.91 & 79.97	 &	81.22\\
         
          \hdashline
         DeBERTa+\textit{PPS} & \textbf{83.70} & \textbf{81.52} &		\textbf{82.56}\\

        \bottomrule
    \end{tabular}}
    \caption{Precision, Recall, and F1 for different methods.}
    \label{tab:test_res}
\end{table}

\section{Discussion}

\subsection{Comparison between DeBERTa and DeBERTa+\textit{PPS}}

\begin{figure}
    % \centering
        \includegraphics[width=1.02\columnwidth]{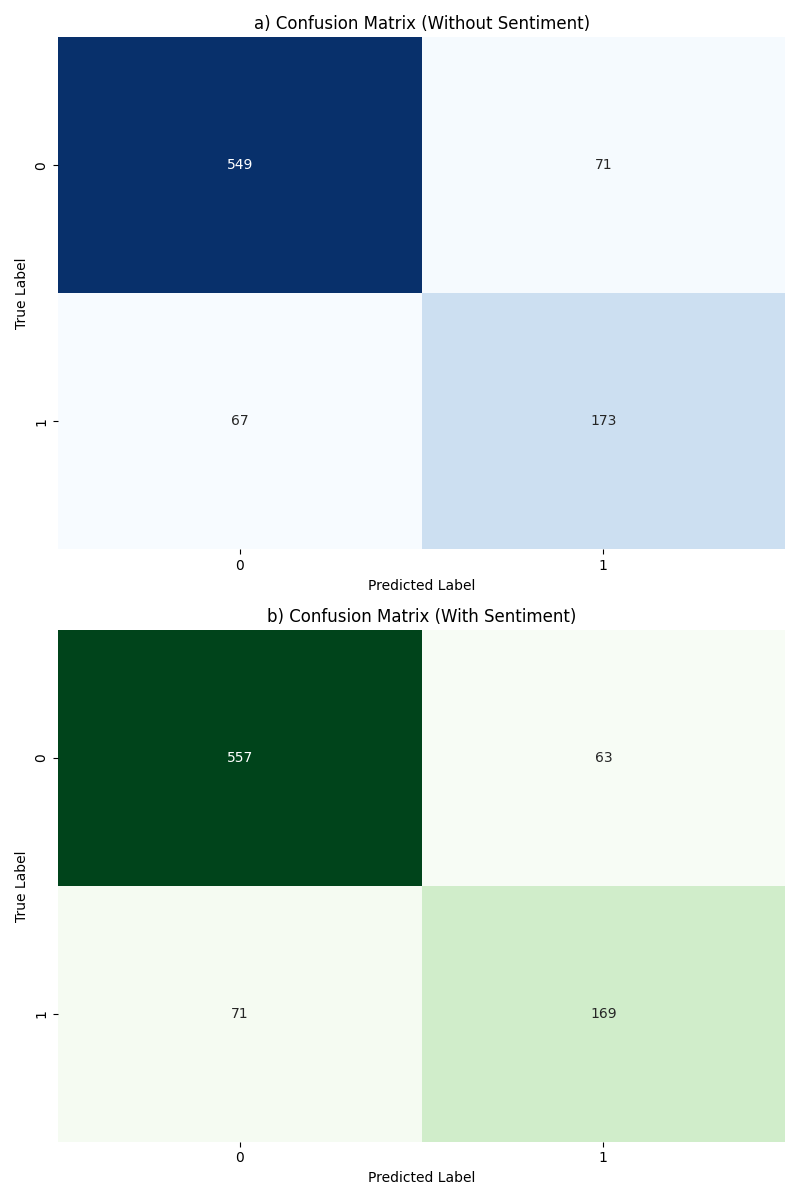}
    \caption{Confusion Matrix of a) DeBERTa and b) DeBERTa+\textit{PPS} depicting effect of sentiment on the DeBERTa model.}
    \label{fig:confusion_matrix_comparing_pps}
\end{figure}

Figure \ref{fig:confusion_matrix_comparing_pps} shows how much effect sentiment had on different OLID labels. Notice that NOT labels have higher true positives with \textit{PPS} compared to that of not having \textit{PPS}. 
On the other hand, the performance decreased on the OFF instances. In fact, we notice DeBERTa+\textit{PPS} predicts overall less OFF compared to that of DeBERTa. As a result, we can conclude that \textit{PPS}-OLID only improves model on not predict the OFF instances better.

\subsection{Effect of Sentiment on OLID}

\begin{figure}
    % \centering
        \includegraphics[width=1.02\columnwidth]{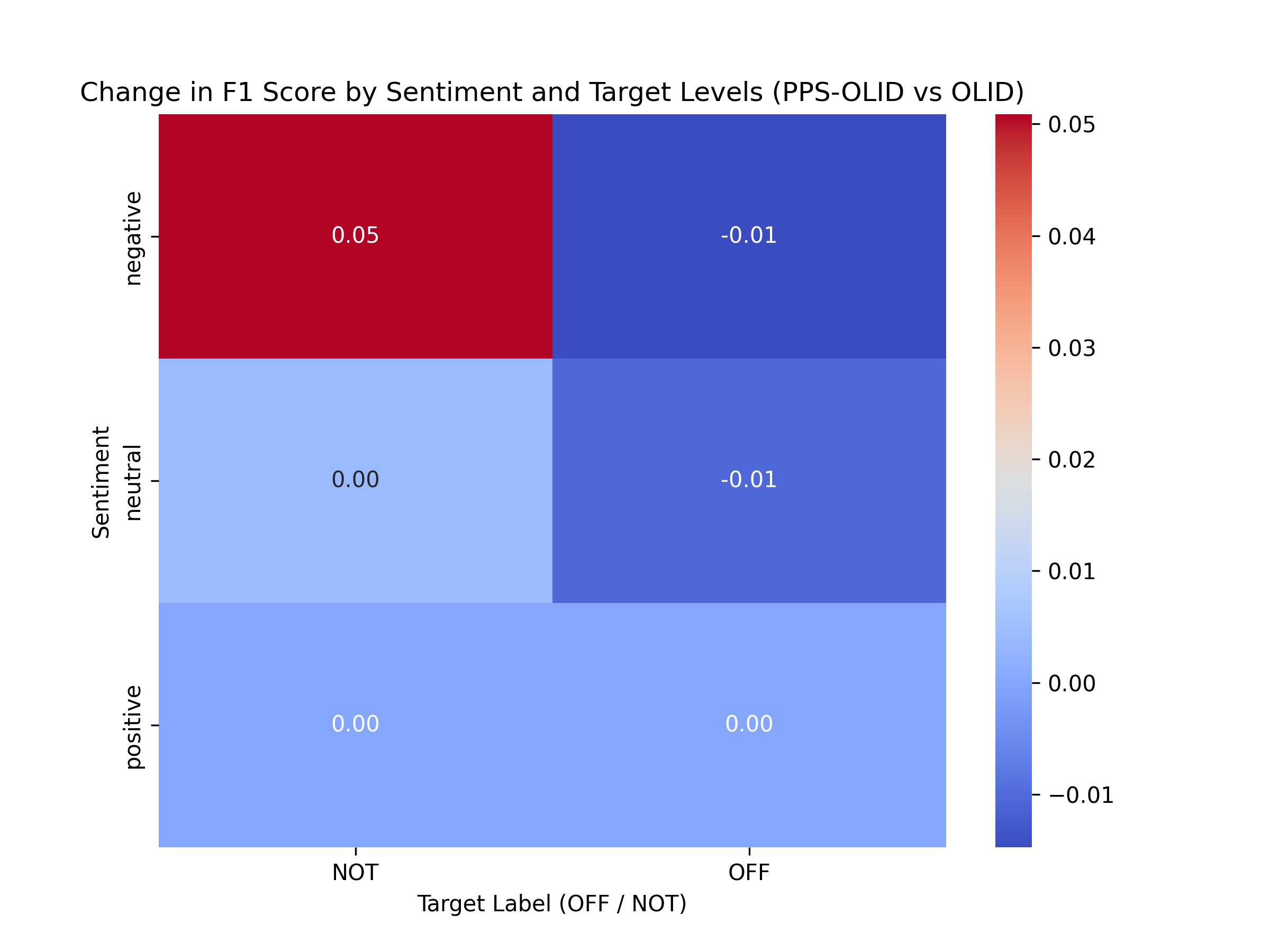}
    \caption{Heatmap on the Affect of Each Sentiment on DeBERTa model before and after \textit{PPS}-OLID.}
    \label{fig:change_in_f1}
\end{figure}

Figure \ref{fig:change_in_f1} further shows which sentiment helped better identify OFF and NOT labels on DeBERTa+\textit{PPS} model compared to the DeBERTa model. Notice that the \textit{Negative} sentiment helps better predict the NOT instances instead of the expected \textit{Neutral} sentiment. In fact, the \textit{Neutral} sentiment had no positive effect. This is surprising as \newcite{sharma2021deep} probed into their hypothesis believing that \textit{Neutral} sentiment has a high correlation with NOT label. 

Also, notice that the \textit{Positive} sentiment had no impact on any of the OLID labels. One reason could be due to the small sample of \textit{Positive} sentiment predicted on this dataset. Recall, this is also true for \textit{Negative} sentiment compared to that of \textit{Neutral} sentiment. As a result, future work should revolve around utilizing transfer learning which will enable a more diverse set of samples to continue examining the underlying effect of sentiment on OLID tasks.

\section{Conclusion}
In this work, we experiment whether leveraging sentiment increases the performance of several feature based and deep learning models on offensive language classification task. We augment the OLID dataset by prepending the sentiment predicted by a pretrained language models. Later, we compare and analyse the performance of several baseline models on OLID dataset. Later, we pick the best performing, DeBERTa, model and train with our augmented dataset to monitor whether the model performs better. We find that leveraging sentiment indeed increased the DeBERTa model by $0.28$ F1-score. Later, we perform error analysis on our augmented dataset and suggest that future work should revolve conducting this experiment on a larger dataset.

\typeout{}

\bibliography{bib}
\bibliographystyle{eacl2017}

\clearpage
\appendix
\section*{Appendix}

\section{Hyper-parameters and their Values}\label{app:app_sec_hyper_paramete}

% % Insert Hyper-parameter values for LSTM
\begin{table}[ht]
    \centering
    \resizebox{0.9\columnwidth}{!}{%
    \begin{tabular}{|l|m{10em}|} 
     \hline
     \textbf{Hyper Parameter} & \textbf{Exploration Space}\\
     \hline
     \textbf{Batch Size} & 16, 32*, 64 \\
     \hline
     \textbf{LSTM Units} & 50*, 64, 100*, 128, 256 \\
     \hline
     \textbf{Learning Rate ($\lambda$)} & 1 $\times$ 10$^{-3}$, 1 $\times$ 10$^{-4}$, 5 $\times$ 10$^{-4}$* \\
     \hline
     \textbf{Dropout Rate} & 0.10, 0.20*, 0.30*, 0.40 \\
     \hline
     \textbf{Epochs} & 20, 30, 50*. 70 \\
     \hline
     \textbf{Layers} & 1, 2, 3* \\
     \hline
    \end{tabular}}
    \caption{Hyper-parameters and their values for LSTM models. * indicates the best-performing value.}
    \label{tab:Hyper_Parameter_LSTM_Tab}
\end{table}

% Insert Hyper-parameter values for Transformer
\begin{table}[ht]
    \centering
    \resizebox{0.9\columnwidth}{!}{%
    \begin{tabular}{|l|m{10em}|} 
     \hline
     \textbf{Hyper Parameter} & \textbf{Exploration Space}\\
     \hline
     \textbf{Batch Size} & 8, 16*, 32 \\
     \hline
     \textbf{Weight Decay} & 1 $\times$ 10$^{-2}$, 1 $\times$ 10$^{-1}$, 2 $\times$ 10$^{-1}$, 3 $\times$ 10$^{-1}$*, 4 $\times$ 10$^{-1}$, 5 $\times$ 10$^{-1}$, \\
     \hline
     \textbf{Learning Rate} & 2 $\times$ 10$^{-5}$*, 2 $\times$ 10$^{-4}$ \\
     \hline
     \textbf{Epochs} &  3, 4*, 5 \\
     \hline
    \end{tabular}}
    \caption{Hyper-parameters and their values for Transformer models. * indicates the best-performing value.}
    \label{tab:Hyper_Parameter_Transformer_Tab}
\end{table}

\section{Results for Each Class}\label{app:tab_class}

% % Insert results for each class of the DeBERTa model
\begin{table}[ht]
    \centering
    \resizebox{0.99\columnwidth}{!}{%
    \begin{tabular}{|l|c|c|c|}
        \hline
        \textbf{Classes} & \textbf{Precision} & \textbf{Recall} & \textbf{F1} \\
        \hline
        NOT & 70.90 & 72.08 & 71.49\\
        OFF & 89.12 & 88.55 & 88.83\\
        \hline
    \end{tabular}}
    \caption{Precision, Recall, and F1 for each class on DeBERTa.}
    \label{tab:per_class_metric}
\end{table}

\section{Complete result of DeBERTa}\label{app:complete_set}

\begin{table*}[t]
    \centering
    \resizebox{1.8\columnwidth}{!}{%
    \begin{tabular}{c c c c c c c c}
        \toprule
        \textbf{Batch Size} & \textbf{Weight Decay} &\textbf{Lr} & \textbf{Max Length} & \textbf{Precision} &\textbf{Recall} & \textbf{F1} \\
        \midrule
        16 &	0.01 &	2 $\times$ 10$^{-5}$ &	128 &	79.63 &	79.24 &	79.43	 \\
        16 &	0.01 &	2 $\times$ 10$^{-5}$ &	50 &	40.00 &	40.00 &	40.00   \\
        16 &	0.1 &	2 $\times$ 10$^{-5}$ &	128 &	79.04 &	81.83 &	80.14	  \\
        16&	0.1&	2 $\times$ 10$^{-4}$&	128&	40.00&	40.00&	40.00 \\
        32&	0.1&	2 $\times$ 10$^{-5}$&	128&	82.15&	77.58&	79.34 \\
         
        16&	0.1&	2 $\times$ 10$^{-5}$&	128&	40.00&	40.00&	40.00 \\
         16&	0.2&	2 $\times$ 10$^{-5}$&	128&	80.87&	80.87&	80.87 \\
         16&	0.5&	2 $\times$ 10$^{-5}$&	128&	79.50&	79.03&	79.26 \\
         16&	0.3&	2 $\times$ 10$^{-5}$&	128&	80.81&	81.13&	80.97	 \\
       16&	0.3&	2 $\times$ 10$^{-5}$&	128&	78.90&	79.69&	79.27\\
         16&	0.4&	2 $\times$ 10$^{-5}$&	128&	77.90&	79.39&	78.56 \\
         16&	0.3&	2 $\times$ 10$^{-5}$&	128&	83.12&	77.09&	79.26 \\
         16&	0.3&	2 $\times$ 10$^{-5}$&	75&	81.96&	82.29&	82.12 \\
        16&	0.3&	2 $\times$ 10$^{-5}$&	75&	82.94&	81.71&	82.28\\
        
         16&	0.3&	2 $\times$ 10$^{-5}$&	76&	\textbf{83.70}&	\textbf{81.52}&	\textbf{82.56}  \\
        
        \bottomrule
    \end{tabular}}
    \caption{Performance of DeBERTa on range of different hyper parameters.}
    \label{tab:complete_test_res}
\end{table*}

\end{document}